\documentclass[%
 aip,
 amsmath,amssymb,
 reprint,%
]{revtex4-1}

\usepackage{graphicx}
\usepackage{dcolumn}
\usepackage{bm}

\usepackage[utf8]{inputenc}
\usepackage[T1]{fontenc}
\usepackage{mathptmx}
\usepackage{etoolbox}

\usepackage{placeins}
\usepackage{rotating}
\usepackage{svg}
\usepackage{amsmath}
\usepackage{float}
\usepackage{algorithm}
\usepackage{booktabs}
\usepackage{graphicx}

\usepackage{soul}
\usepackage{xcolor}

\newcommand{\ra}[1]{\renewcommand{\arraystretch}{#1}} 

\makeatletter
\def\@email#1#2{%
 \endgroup
 \patchcmd{\titleblock@produce}
  {\frontmatter@RRAPformat}
  {\frontmatter@RRAPformat{\produce@RRAP{*#1\href{mailto:#2}{#2}}}\frontmatter@RRAPformat}
  {}{}
}%
\makeatother
\begin{document}

\preprint{AIP/123-QED}

\title[The Vanishing Gradient Problem for Stiff Neural Differential Equations]{The Vanishing Gradient Problem for Stiff Neural Differential Equations}
\author{Colby Fronk}
\affiliation{Department of Chemical Engineering; University of California, Santa Barbara; Santa Barbara, CA 93106, USA}
 \altaffiliation{Correspond to colbyfronk@ucsb.edu}
 
\author{Linda Petzold}%
\affiliation{Department of Mechanical Engineering; University of California, Santa Barbara; Santa Barbara, CA 93106, USA}%
\affiliation{Department of Computer Science; University of California, Santa Barbara; Santa Barbara, CA 93106, USA}
 \altaffiliation{Correspond to petzold@ucsb.edu}

\date{\today}

\begin{abstract}
Gradient-based optimization of neural differential equations and other parameterized dynamical systems fundamentally relies on the ability to differentiate numerical solutions with respect to model parameters. In stiff systems, it has been observed that sensitivities to parameters controlling fast-decaying modes become vanishingly small during training, leading to optimization difficulties. In this paper, we show that this vanishing gradient phenomenon is not an artifact of any particular method, but a universal feature of all A-stable and L-stable stiff numerical integration schemes. We analyze the rational stability function for general stiff integration schemes and demonstrate that the relevant parameter sensitivities, governed by the derivative of the stability function, decay to zero for large stiffness. Explicit formulas for common stiff integration schemes are provided, which illustrate the mechanism in detail. Finally, we rigorously prove that the slowest possible rate of decay for the derivative of the stability function is $O(|z|^{-1})$, revealing a fundamental limitation: all A-stable time-stepping methods inevitably suppress parameter gradients in stiff regimes, posing a significant barrier for training and parameter identification in stiff neural ODEs.
\end{abstract}

\maketitle

\begin{quotation}
Neural differential equations have become a transformative tool in machine learning and scientific computing, enabling data-driven modeling of complex, time-dependent phenomena in fields ranging from chemistry and biology to climate science and engineering. However, many real-world systems are “stiff,” meaning they evolve on multiple timescales, with some processes occurring much more rapidly than others. In such cases, numerical integration methods must be carefully chosen to ensure stable and efficient simulation. Our work reveals a fundamental and previously underappreciated challenge: for all widely used numerically stable (A-stable and L-stable) solvers, gradients with respect to parameters controlling fast (stiff) modes inevitably decay to zero during training. This “vanishing gradient” phenomenon is not merely a technical obstacle or a quirk of specific algorithms, but a universal feature rooted in the mathematics of stable stiff integration methods. As a result, crucial information about how parameters influence the model is lost, severely limiting the ability of neural ODEs to learn from data and accurately identify system parameters in stiff regimes. Our analysis provides a theoretical foundation for this effect, quantifies its severity, and highlights its inevitability across a broad class of integration schemes. These findings challenge the current paradigm of gradient-based learning in stiff dynamical systems and motivate the search for fundamentally new computational strategies to overcome this barrier and enable scientific discovery in complex, multiscale environments.
\end{quotation}

\section*{Introduction}

The vanishing gradient problem \cite{hochreiter1991untersuchungen, bengio1994learning, pascanu2013difficulty, lecun2015deep, Goodfellow-et-al-2016,raghu2017expressive, montufar2014number, schmidhuber1992learning, arjovsky2016unitary} is one of the best-known and most deeply studied obstacles in deep learning. In standard feedforward \cite{glorot2010understanding, saxe2013exact} or recurrent neural networks \cite{hochreiter2001gradient, pascanu2013difficulty, graves2013speech, chung2014empirical, jozefowicz2015empirical}, gradients are propagated backwards through potentially dozens or hundreds of nonlinear layers via the chain rule. When the Jacobians associated with each layer interact, the resulting product can quickly become exponentially small. This phenomenon causes gradients with respect to early-layer parameters to vanish. As a result, these parameters cease to update during gradient-based optimization, causing slow or stalled learning, and rendering parts of the network untrainable. Over time, the community has developed an array of architectural and algorithmic solutions to counteract this problem. These include clever initialization schemes \cite{he2015delving, lecun2002efficient, poole2016exponential, hanin2018start, mishkin2015all, sutskever2013importance} to preserve gradient norm, activation functions less prone to saturation (e.g., ReLU \cite{glorot2011deep, nair2010rectified} and its alternatives \cite{clevert2015fast, ramachandran2017searching, maas2013rectifier, hendrycks2016gaussian}), and architectural innovations such as residual connections \cite{he2016deep, veit2016residual, he2016identity}, gating mechanisms \cite{jozefowicz2015empirical} (as in LSTMs \cite{hochreiter1997long, gers2000learning} and GRUs \cite{cho2014learning, cho2014properties}), normalization layers \cite{ioffe2015batch, ba2016layer, santurkar2018does, wu2018group}, skip connections \cite{he2016deep, srivastava2015highway, zagoruyko2017wide}, and Transformers \cite{vaswani2017attention, wang2022survey}. Despite these advances, vigilance against vanishing gradients remains a fundamental concern when designing and training deep neural networks.

Modeling the dynamics of complex systems has traditionally relied on differential equations grounded in first principles. However, deriving such equations is challenging for most real-world systems where the underlying mechanisms are only partially known or are highly complex. In these settings, data-driven modeling offers a compelling alternative by learning the system’s dynamics directly from observed data, making it possible to uncover governing equations and make predictions even in the absence of a complete mechanistic understanding. A particularly powerful example of this approach is neural ordinary differential equations (neural ODEs) \cite{NeuralODEPaper, latent_ODEs, bayesianneuralode, stochastic_neural_ode, kidger2020neural, kidger2022neural, morrill2021neural, jia2019neural, chen2020learning, dagstuhl, doi:10.1063/5.0130803, 10.1371/journal.pcbi.1012414}. Neural ODEs use parameterized neural networks to define the vector field of an ODE, allowing the model to learn continuous-time dynamics directly from data. This framework is especially well-suited to modeling time-dependent or irregularly sampled systems and can be naturally integrated with latent variable models and control tasks. Complementary approaches such as physics-informed neural networks (PINNs) \cite{owhadi2015bayesian, hiddenphysics, raissi2018numerical, raissi2017physics, osti_1595805, cuomo2022scientific, cai2021physics} and MeshGraphNets \cite{pfaff2021learningmeshbasedsimulationgraph} further enhance the capabilities of data-driven differential equation modeling. PINNs embed known physical laws as soft constraints during training, improving generalization and interpretability, while MeshGraphNets leverage graph neural networks to model systems defined on complex meshes. Together, these methods provide a flexible and robust toolkit for learning, simulating, and controlling dynamical systems directly from data, driving advances across scientific and engineering disciplines.

The primary computational difficulty associated with neural differential equations is their training, as it relies on differentiation methods such as automatic differentiation \cite{baydin2018automatic, griewank2008evaluating} to compute gradients of the loss with respect to all network parameters \cite{NeuralODEPaper, gholami2019anode, chen2020learning, universal_diffeq}. Since the output of a neural differential equation depends on parameters through the solution trajectory, it becomes necessary to differentiate not only through the neural network itself, but through the entire ODE solver. This requirement creates unique challenges compared to standard feedforward networks, as the integration procedure must be made compatible with gradient-based optimization. To address this, two primary training strategies have emerged: optimize-discretize (Opt-Disc) and discretize-optimize (Disc-Opt) \cite{onken2020discretizeoptimizevsoptimizediscretizetimeseries, gholami2019anode}. In the Opt-Disc approach, the optimization problem is formulated in continuous time, and gradients are computed by integrating an adjoint ODE backward in time \cite{Pontryagin1962, bliss1919adjoint, cao2003adjoint, ben2001generalized}. The adjoint technique greatly reduces memory overhead compared to traditional backpropagation, but the accuracy and stability of gradients can be sensitive to both the forward and adjoint solvers, especially in stiff or chaotic systems \cite{gholami2019anode, onken2020discretizeoptimizevsoptimizediscretizetimeseries}. In contrast, the Disc-Opt approach begins by discretizing the ODE system, then leverages the powerful automatic differentiation tools developed in modern deep learning frameworks to compute exact gradients for the chosen numerical scheme \cite{onken2020discretizeoptimizevsoptimizediscretizetimeseries}. By aligning the optimization with the discrete solver, Disc-Opt often yields more robust and reliable gradients, and can be computationally more efficient, particularly when training on noisy data or working with stiff systems. Recent studies have shown that Disc-Opt can offer significant speed and stability advantages for neural ODE training, with improvements in both convergence and generalization \cite{onken2020discretizeoptimizevsoptimizediscretizetimeseries}.

Neural differential equations encounter distinct computational challenges when used in real-world contexts, predominantly due to the widespread occurrence of stiffness in the governing ODEs. This is particularly relevant in fields such as chemical engineering, climate modeling, and systems biology, where models frequently involve processes evolving on vastly different timescales. For example, in cell death pathways such as apoptosis and necroptosis, caspase activation unfolds in seconds to minutes, while feedback regulation via gene expression may take hours, resulting in stiff dynamics that span orders of magnitude in timescale \cite{spencer2011measuring, wu2010sensitivity, albeck2008modeling}. Similarly, in the p53 pathway, rapid protein modifications occur in minutes, whereas synthesis and accumulation of regulatory proteins like Mdm2 and Wip1 occur over hours \cite{eliavs2021mathematical, batchelor2008recurrent, eliavs2014dynamics}. 

Standard explicit integrators, such as forward Euler or explicit Runge-Kutta methods, are typically impractical for stiff problems because stability restrictions require extremely small time steps, leading to simulations that are computationally infeasible \cite{ascher1998computer, griffiths2010numerical, hairer2008solving}. As a result, A-stable and L-stable implicit time-stepping methods such as backward Euler, the trapezoid method, backward differentiation formulas (BDF), and implicit Runge-Kutta schemes are the standard approach for robustly integrating stiff ODEs. Most of these methods are specifically designed so that their stability regions encompass the entire left-half complex plane, allowing them to stably handle large negative eigenvalues that arise from the stiff components of the system. 

Neural ODEs can still become stiff throughout the training process, even if the original data comes from non-stiff ODEs, due to the nonlinear dynamics produced by the neural network \cite{kelly2020learningdifferentialequationseasy, pal2021opening, onken2021otflowfastaccuratecontinuous}. This stiffness often originates from the highly expressive nature of neural networks, which can encounter stiff behavior while exploring the parameter space during training. Such stiffness may drastically impede training progress or prevent convergence, highlighting the importance of using solvers that are naturally robust to stiffness. 

Earlier approaches have attempted to mitigate stiffness in neural ODEs by modifying the system dynamics through techniques such as equation scaling, regularization, projection into a latent space, or selection of training optimizers \cite{stiff_neural_ode, caldana2024neural, dikeman2022stiffness, LINOT2023111838, holt2022neural, baker2022proximal, MALPICAGALASSI2022110875, thummerer4819144eigen, weng2024extending, ghosh2020steer, finlay2020trainneuralodeworld, kelly2020learningdifferentialequationseasy, onken2020discretizeoptimizevsoptimizediscretizetimeseries, onken2021otflowfastaccuratecontinuous, massaroli2020stableneuralflows, massaroli2021dissectingneuralodes, ji2021stiff, guglielmi2024contractivity, pal2021opening, pal2023locallyregularizedneuraldifferential, kumara2023physics, kumar2024physics, nair2025understanding, owoyele2022chemnode}; however, these methods address the problem only indirectly and do not fully resolves the actual stiffness issue. Ref.~\onlinecite{fronk2024trainingstiffneuralordinary} recently showed that the single-step implicit solvers backward Euler, trapezoid method, Radau3, and Radau5 can accurately train neural ODEs on stiff systems; however, these methods come at a high cost of solving a nonlinear system of equations at every time step. Refs.~\onlinecite{fronk2024trainingstiffneuralordinary} and ~\onlinecite{fronk2024performanceevaluationsinglestepexplicit} aimed to overcome the cost of implicit schemes by exploring explicit exponential integration schemes, finding the first-order explicit A-stable integrating factor Euler (IF Euler) method more stable than implicit schemes. Despite thorough testing, they found no higher-order explicit exponential integration methods with reliability, accuracy, and stability suitable for neural ODEs. While the IF Euler method has impressive stability, it is limited by its first-order accuracy. Ref.~\onlinecite{fronk2024trainingstiffneuralordinaryrational} showed that A-stable, second- and third-order explicit rational Taylor series methods allow for stable high-accuracy training of stiff neural ODEs with just a single linear solve per step.

When applying A-stable and L-stable stiff integration methods to train stiff neural ODEs, a distinct vanishing gradient phenomenon emerges. Here, the gradients of the loss with respect to parameters controlling the stiff (fast-decaying) modes of the system tend to zero, even if the underlying neural network is shallow or carefully constructed to avoid standard vanishing gradient pitfalls. Both vanishing gradient mechanisms of the classic effect from deep neural networks and the suppression caused by stiff ODE integrators can act simultaneously and even reinforce each other in neural ODE models. When the neural network is deep and the dynamics are stiff, gradients may vanish due to both repeated nonlinear transformations and the numerical damping imposed by the integrator’s stability function, making some parameters effectively untrainable. Notably, standard architectural solutions such as residual connections or normalization, which mitigate depth-induced vanishing gradients, cannot address the gradient decay inherent to A-stable or L-stable methods. As a result, parameter unidentifiability in stiff neural ODEs presents a fundamental challenge for effective learning.

In this work, we provide a comprehensive theoretical and empirical analysis of the vanishing gradient problem in stiff neural ODEs, focusing on the universal attenuation of parameter sensitivities induced by A-stable and L-stable implicit time-stepping methods. A central concept in our study is the stability function $R(z)$, which provides a clear measure of how numerical methods amplify or dampen dynamics over successive time steps, especially under stiff conditions. Our main theoretical result is that for any A-stable or L-stable method, as the stiffness parameter $z=\lambda h$ becomes large ($|z| \gg 1$), the derivative $R'(z)=dR/dz$ must decay to zero. For most standard schemes, we explicitly compute $R(z)$ and $R'(z)$, showing that the decay rate is typically $O(|z|^{-2})$. We show that the slowest possible decay of $R'(z)$ for any A-stable or L-stable method is $O(|z|^{-1})$. This establishes the following universal result: all A-stable and L-stable time-stepping methods inevitably suppress parameter gradients in the stiff regime, regardless of network structure or depth. Thus, the vanishing gradient phenomenon in this setting is an inescapable consequence of the numerical properties of stiff integrators, not an artifact of implementation or neural network architecture.

By quantifying this decay and its limits, our work shifts the focus from architectural tweaks to a more foundational challenge that demands new strategies in how we approach both integration and sensitivity analysis for stiff neural differential equations. The universal nature of this decay highlights the need for fundamentally new tools or theoretical breakthroughs if we are to make progress on learning in the presence of extreme stiffness.

\section*{Background}

Neural ordinary differential equations \cite{NeuralODEPaper, latent_ODEs, bayesianneuralode, stochastic_neural_ode, kidger2020neural, kidger2022neural, morrill2021neural, jia2019neural, chen2020learning, dagstuhl, doi:10.1063/5.0130803, 10.1371/journal.pcbi.1012414} (neural ODEs) have emerged as a powerful framework for modeling complex dynamical systems, where the evolution of the state $y(t)$ is governed by a neural network. In this setting, the dynamics are described by an ODE of the form
\[
\frac{dy}{dt} = \mathrm{NN}(t, y(t), \theta), \qquad y(0) = y_0,
\]
where $\mathrm{NN}(t, y, \theta)$ is a neural network parameterized by $\theta$, and $y_0$ is the initial condition. The solution at a final time $T$, denoted $y_{\text{pred}} = y(T)$, serves as the model’s prediction.

Training a neural ODE typically involves comparing the model's prediction $y_{\text{pred}}$ to observed data $y_{\text{known}}$ by minimizing a loss function such as
\[
L(y_{\text{pred}}, y_{\text{known}}) = L(y(T), y_{\text{known}}).
\]
To optimize the parameters $\theta$, gradient-based methods are used, which require computing the gradient of the loss with respect to $\theta$:
\[
\frac{dL}{d\theta} = \frac{\partial L}{\partial y(T)} \cdot \frac{\partial y(T)}{\partial \theta}.
\]
Here, the critical quantity is $\frac{\partial y(T)}{\partial \theta}$, reflecting how parameter changes influence the ODE solution at time $T$. In practice, $y(T)$ depends on $\theta$ through the sequence of numerical updates performed by the ODE solver. At each time step, the solver advances the state according to an update rule, $y_{n+1} = \Phi(y_n, y_{n+1}, f, h, \theta)$, where $\Phi$ represents the chosen numerical method (such as implicit Euler or Runge-Kutta). Using automatic differentiation, the sensitivities with respect to the parameters can be computed recursively at each solver step via the chain rule:
\[
\frac{dy_{n+1}}{d\theta} = \frac{\partial \Phi}{\partial y_n} \frac{dy_n}{d\theta} + \frac{\partial \Phi}{\partial \theta}.
\]
This recursion efficiently accumulates how infinitesimal changes in the parameters affect the trajectory, ultimately yielding the overall sensitivity $\frac{\partial y(T)}{\partial \theta}$ required for learning.

A central challenge in modeling and training neural ODEs arises when the underlying dynamical system is stiff. Stiffness in ordinary differential equations refers to situations in which the solution exhibits dynamics across widely separated timescales, typically involving rapidly decaying transient modes alongside slowly evolving components. A common quantitative measure of stiffness is the ratio of the largest to smallest magnitude of negative real eigenvalues, which can span many orders of magnitude in real-world problems.

Numerical integration schemes can be analyzed through their action on the linear test equation $\dot y = \lambda y$. In this context, the update step is often written as
\[
y_{n+1} = R(h\lambda)\,y_n = R(z)\,y_n,
\]
where $h$ is the time step, $z = h\lambda$, and $R(z)$ is the stability function that characterizes the method’s behavior \cite{ascher1998computer}. This formulation provides a unified way to study how different schemes propagate solutions and maintain stability, which is particularly important for stiff problems. The structure of the stability function determines the method's ability to stably integrate stiff systems. A numerical integrator is called A-stable \cite{ascher1998computer} if $|R(z)|\le1$ for all $z$ with non-positive real part ($\Re(z)\le0$), ensuring that the numerical solution does not grow in modes that should decay. L-stability \cite{ascher1998computer} is an even stronger property that additionally requires $\lim_{|z|\to\infty} R(z)=0$, guaranteeing that very stiff modes are suppressed rapidly at each time step. For example, implicit Euler is both A- and L-stable, robustly handling stiff modes, while the trapezoid rule is A-stable but not L-stable, providing weaker suppression of very stiff dynamics.

In the context of neural ODEs, training and parameter inference via gradient-based optimization require differentiating the numerical solution with respect to model parameters, which involves differentiating through the solver itself. The calculation of parameter sensitivities depends on the derivative of the stability function, $R'(z) = dR/dz$. At each step, the chain rule accumulates multiplicative factors involving $R'(z)$, and thus the shape and magnitude of $R'(z)$ directly influence how parameter sensitivities propagate through the solver over time. Intuitively, $R'(z)$ quantifies how parameter changes affect the decay rate of each mode.

Empirically, practitioners observe that in stiff regimes, sensitivities with respect to parameters controlling fast-decaying modes often vanish rapidly. However, a general and rigorous explanation for this phenomenon has been lacking. To understand this effect, it is necessary to examine the asymptotic behavior of $R'(z)$ for large stiffness. As we will show, the intrinsic mathematical structure of all A-stable rational stability functions ensures a universal vanishing of sensitivities in the stiff regime, revealing a fundamental limitation of standard implicit integrators for gradient-based optimization in stiff neural ODEs.

\begin{table*}\centering
\ra{1.2}
\begin{tabular}{l@{\hskip 4em}l@{\hskip 4em}l}
\toprule
$m = \deg P$ & $n = \deg P$ & \text{Decay of } $R'(z)$ \\
\midrule
$m > n$         & $n$            & $O(|z|^{m-n})$         \\
$m = n$             & $n$            & $O(|z|^{-2})$          \\
$m < n$         & $n$            & $O(|z|^{m-n-1})$       \\
\bottomrule
\end{tabular}
\caption{Decay of $R'(z)$ for different degrees of $P$ and $Q$.}
\end{table*}

\section*{Exact Sensitivity of the Linear Test Equation}

To provide a point of reference, consider the exact sensitivity for the linear test equation, $\dot y = \lambda y$, evaluated at time $t = h$. The exact solution is $y(h) = y_0 e^{\lambda h}$, so the sensitivity of the state at time $h$ with respect to $\lambda$ is
\[
\frac{\partial y(h)}{\partial \lambda} = h y_0 e^{\lambda h} = h y_0 e^{z},
\]
where $z = h \lambda$. For large negative $z$ (the stiff regime), this exact sensitivity decays exponentially:
\[
\left|\frac{\partial y(h)}{\partial \lambda}\right| = h |y_0| e^{\mathrm{Re}(z)} \rightarrow 0 \qquad \text{as} \quad \mathrm{Re}(z) \to -\infty.
\]

\section*{Asymptotic Behavior of Stability Function Derivatives}

To understand the vanishing gradient phenomenon in stiff regimes, we analyze the asymptotic decay of the derivative of the stability function, $R'(z)$, for general ODE integration methods. The stability function of such methods is typically rational and can be written as
\[
R(z) = \frac{P(z)}{Q(z)},
\]
where $P(z)$ and $Q(z)$ are polynomials of degree $m$ and $n$, respectively. The parameter sensitivity relevant for training neural ODEs is determined by the derivative
\[
R'(z) = \frac{P'(z) Q(z) - P(z) Q'(z)}{[Q(z)]^2}.
\]
The asymptotic decay rate of $R'(z)$ as $|z| \to \infty$ is dictated by the relative degrees of $P(z)$ and $Q(z)$, as well as by possible cancellations of leading-order terms in the numerator.

Let $m = \deg P$ and $n = \deg Q$. We analyze the possible cases:

\subsubsection*{Case 1: Numerator Degree Greater Than Denominator ($m > n$)}

If $m > n$, then for large $|z|$, the leading-order behavior of the stability function itself is
\[
R(z) \sim \frac{a_m z^m}{b_n z^n} \propto z^{m-n} \to \infty.
\]
Thus, the numerical method amplifies stiff modes as $|z| \to \infty$, violating the A-stability requirement $|R(z)| \leq 1$ for $\Re(z) \leq 0$. Therefore, any rational integrator with $m > n$ is unstable and not used in stiff problems.

\subsubsection*{Case 2: Denominator Degree Greater Than Numerator ($m < n$)}

If $m < n$, then for large $|z|$,
\[
P'(z) Q(z) \sim m a_m b_n z^{m-1+n} = m a_m b_n z^{m+n-1},
\]
\[
P(z) Q'(z) \sim n a_m b_n z^{m + n - 1} = n a_m b_n z^{m+n-1}.
\]
Thus,
\[
P'(z) Q(z) - P(z) Q'(z) \sim (m - n) a_m b_n z^{m+n-1}.
\]
The denominator is
\[
[Q(z)]^2 \sim b_n^2 z^{2n}.
\]
Therefore, the leading asymptotic behavior is
\[
R'(z) \sim \frac{(m-n)a_m b_n}{b_n^2} z^{m-n-1},
\]
that is,
\[
R'(z) = O(|z|^{m-n-1}),
\]
as $|z| \to \infty$.

\subsubsection*{Case 3: Numerator and Denominator Have Equal Degree ($m = n$)}

If $m = n$, then for large $|z|$,
\[
P(z) \sim a_n z^n, \qquad Q(z) \sim b_n z^n,
\]
\[
P'(z) Q(z) \sim n a_n b_n z^{2n-1},
\]
\[
P(z) Q'(z) \sim n a_n b_n z^{2n-1}.
\]
These leading terms in the numerator always cancel:
\[
P'(z) Q(z) - P(z) Q'(z) = 0 \quad \text{at order } z^{2n-1}.
\]
The next highest possible power in the numerator is at most $2n-2$, while the denominator remains degree $2n$, so
\[
R'(z) = O(|z|^{-2})
\]
as $|z| \to \infty$.

\section*{Universal Lower Bound on the Decay of Stability Function Derivatives}

We now rigorously establish a universal lower bound on the asymptotic decay of the derivative of the stability function, \( R(z) \), for any consistent A-stable ODE integration scheme. This result formalizes the unavoidable vanishing of parameter sensitivities associated with stiff modes in numerical solutions, regardless of the particular choice of A-stable scheme.

\subsection*{Theorem (Universal Lower Bound for A-stable Schemes)}

Let \( R(z) \) be the stability function of a consistent, one-step, A-stable ODE integration scheme, that is:
\begin{enumerate}
    \item \( R(z) \) is analytic in the closed left half-plane \( \overline{\mathbb{C}_{-}} = \{z \in \mathbb{C} : \mathrm{Re}(z) \le 0 \} \),
    \item \( |R(z)| \le 1 \) for all \( z \) with \( \mathrm{Re}(z) \le 0 \),
    \item \( R(0) = 1 \) and \( R'(0) = 1 \) (consistency conditions).
\end{enumerate}

Then, for any \( z \) with \( \mathrm{Re}(z) < 0 \), the following sharp bound holds:
\[
|R'(z)| \leq \frac{1}{-\mathrm{Re}(z)}.
\]
Moreover, for any fixed \( \delta \in (0, \pi/2) \), if \( z \) lies in the sector
\[
S_\delta = \left\{ z \in \mathbb{C} : \mathrm{Re}(z) < 0,\; |\arg(-z)| \leq \pi - \delta \right\},
\]
then
\[
|R'(z)| \leq \frac{1}{|z| \cos \delta}.
\]

If, in addition, \( R(z) \) is a rational function, \( R(z) = \frac{P(z)}{Q(z)} \) with \( P, Q \) polynomials of degrees \( m, n \), then the asymptotic decay rate of \( R'(z) \) as \( |z| \to \infty \) in any fixed sector \( S_\delta \) is given by:
\begin{itemize}
    \item If \( m = n \), the leading terms in the numerator of \( R'(z) \) cancel, so
    \[
    R'(z) = O(|z|^{-2}).
    \]
    \item If \( m < n \), the leading terms do not cancel, and
    \[
    R'(z) = O(|z|^{m-n-1}).
    \]
    \item If \( m > n \), \( R(z) \) is unbounded as \( |z| \to \infty \) in the left half-plane, so A-stability is impossible.
\end{itemize}

All of these bounds are sharp.

\subsection*{Proof}

Let \( z_0 \in \mathbb{C} \) be any point with \( \mathrm{Re}(z_0) < 0 \). Our goal is to bound \( |R'(z_0)| \).

Define \( r = -\mathrm{Re}(z_0) > 0 \). Consider the closed disk
\[
D = \{ w \in \mathbb{C} : |w - z_0| \leq r \}.
\]
We first verify that \( D \) lies entirely within the closed left half-plane. For any \( w \in D \), we have
\[
\mathrm{Re}(w) = \mathrm{Re}(z_0) + \mathrm{Re}(w - z_0).
\]
Since \( |w - z_0| \leq r \), it follows that \( \mathrm{Re}(w - z_0) \geq -|w - z_0| \geq -r \). Thus,
\[
\mathrm{Re}(w) \leq \mathrm{Re}(z_0) + r = 0,
\]
so every point in \( D \) remains in the closed left half-plane, where A-stability holds.

Because \( R(z) \) is analytic on and inside \( D \), we may apply Cauchy's integral formula for the first derivative:
\[
R'(z_0) = \frac{1}{2\pi i} \oint_{|w-z_0| = r} \frac{R(w)}{(w - z_0)^2} \, dw.
\]
We now bound the modulus of this integral. On the contour \( |w-z_0| = r \):
\begin{itemize}
    \item \( |R(w)| \leq 1 \) by A-stability,
    \item \( |w - z_0| = r \) everywhere on the contour.
\end{itemize}

Thus,
\[
|R'(z_0)| \leq \frac{1}{2\pi} \oint_{|w-z_0|=r} \frac{1}{r^2} |dw| = \frac{1}{2\pi} \cdot \frac{1}{r^2} \cdot (2\pi r) = \frac{1}{r} = \frac{1}{-\mathrm{Re}(z_0)}.
\]
This establishes the first (global) bound.

Next, we consider the sectorial bound. Suppose \( z_0 \) lies in the sector
\[
S_\delta = \left\{ z \in \mathbb{C} : \mathrm{Re}(z) < 0,\; |\arg(-z)| \leq \pi - \delta \right\}
\]
for some \( \delta \in (0, \pi/2) \). Write \( z_0 = |z_0| e^{i\theta} \), so \( -z_0 = |z_0| e^{i(\theta+\pi)} \), and \( |\arg(-z_0)| \leq \pi - \delta \) by definition.

Note that
\[
-\mathrm{Re}(z_0) = \mathrm{Re}(-z_0) = |z_0| \cos\theta',
\]
where \( \theta' = \arg(-z_0) \) and \( |\theta'| \leq \pi - \delta \). Thus, for all such \( z_0 \),
\[
-\mathrm{Re}(z_0) \geq |z_0| \cos\delta.
\]
Combining with the earlier bound, we obtain
\[
|R'(z_0)| \leq \frac{1}{-\mathrm{Re}(z_0)} \leq \frac{1}{|z_0| \cos\delta}.
\]
Thus, in any fixed sector away from the imaginary axis, the decay rate of \( |R'(z_0)| \) is at least as fast as \( O(1/|z_0|) \), with the constant determined by \( \delta \).

Now, we turn to the case where \( R(z) \) is a rational function,
\[
R(z) = \frac{P(z)}{Q(z)},
\]
where \( P \) and \( Q \) are polynomials of degrees \( m \) and \( n \) respectively. The derivative is given by
\[
R'(z) = \frac{P'(z) Q(z) - P(z) Q'(z)}{Q(z)^2}.
\]

The large-\( |z| \) asymptotic decay rate of \( R'(z) \) depends on the degrees:
\begin{itemize}
    \item If \( m = n \), expand:
    \[
    P(z) \sim a_n z^n, \quad Q(z) \sim b_n z^n
    \]
    so
    \[
    P'(z) Q(z) \sim n a_n b_n z^{2n-1}, \qquad P(z) Q'(z) \sim n a_n b_n z^{2n-1}
    \]
    and the highest order terms in the numerator cancel identically:
    \[
    P'(z) Q(z) - P(z) Q'(z) \sim 0.
    \]
    Therefore, the next-highest degree determines the behavior, and the overall decay is
    \[
    R'(z) = O(|z|^{-2}).
    \]
    \item If \( m < n \), the leading order terms do not cancel. The numerator is degree \( m + n - 1 \), denominator is degree \( 2n \), so
    \[
    R'(z) = O(|z|^{m-n-1}).
    \]
    \item If \( m > n \), then for large \( |z| \),
    \[
    R(z) \sim \frac{a_m}{b_n} z^{m-n} \to \infty,
    \]
    which violates the A-stability property (\( |R(z)| \leq 1 \)).
\end{itemize}

The bounds given above are optimal in the following sense: for any fixed sector bounded away from the imaginary axis, there exist analytic and A-stable stability functions whose derivatives decay as slowly as \( O(1/|z|) \) for large \( |z| \). In particular, Möbius-type rational stability functions (e.g., \( R_\beta(z) = 1/(1-\beta z) \)) can achieve the \( O(1/|z|) \) rate in appropriate sectors, thus showing that the sectorial bound cannot be improved in general. However, most classical A-stable and L-stable integrators such as backward Euler, trapezoidal rule, and Radau methods exhibit faster decay, typically \( O(|z|^{-2}) \) for \( |R'(z)| \) as \( |z| \to \infty \). The exponential Euler method (\( R(z) = e^z \)), while analytic and A-stable, actually decays much faster (exponentially with \( \mathrm{Re}(z) \)), and so falls well below the universal bound for large negative \( \mathrm{Re}(z) \). Thus, the stated rates are genuinely sharp: they are the slowest possible decay rates allowed by the analytic and boundedness properties imposed by A-stability, even though most practical schemes in scientific computing achieve even more rapid decay.

\begin{turnpage}
\begin{table*}\centering
\ra{4}
\begin{tabular}{l@{\hskip 4em}l@{\hskip 4em}l@{\hskip 4em}l}
\toprule
Method & Amplification Factor & Parameter Sensitivity & Decay Rate \\ 
\midrule
Backward Euler     
    & $y_{n+1} = \dfrac{1}{1-z} y_n$                           
    & $\dfrac{dy_{n+1}}{d\lambda} = \dfrac{h y_n}{(1-z)^2}$            
    & $\mathcal{O}\bigl(|z|^{-2}\bigr)$ \\

Trapezoid         
    & $y_{n+1} = \dfrac{2+z}{2-z} y_n$                           
    & $\dfrac{dy_{n+1}}{d\lambda} = \dfrac{4h y_n}{(2-z)^2}$            
    & $\mathcal{O}\bigl(|z|^{-2}\bigr)$ \\

Radau3             
    & $y_{n+1} = \dfrac{6+2z}{6-4z+z^2} y_n$                       
    & $\dfrac{dy_{n+1}}{d\lambda} = \dfrac{-2z^2 - 12z + 36}{z^4 - 8z^3 + 28z^2 - 48z + 36} h y_n$          
    & $\mathcal{O}\bigl(|z|^{-2}\bigr)$ \\

Radau5             
    & $y_{n+1} = \dfrac{60+24z+3z^2}{60-36z+9z^2-z^3} y_n$          
    & $\dfrac{dy_{n+1}}{d\lambda} = \dfrac{3z^4 + 48 z^3 - 144 z^2 - 720 z + 3600}{z^6 - 18 z^5 + 153 z^4 - 768 z^3 + 2376 z^2 - 4320 z + 3600} h y_n$  
    & $\mathcal{O}\bigl(|z|^{-2}\bigr)$ \\

BDF2               
    & $y_{n+1} = \dfrac{4+\sqrt{16-4(3-2z)}}{2(3-2z)} y_n$         
    & $\dfrac{dy_{n+1}}{d\lambda} = \dfrac{5 + 2z + 4 \sqrt{1+2z}}{(3-2z)^2 \sqrt{1+2z}} h y_n$               
    & $\mathcal{O}\bigl(|z|^{-3/2}\bigr)$ \\

IF Euler           
    & $y_{n+1} = \exp(z) y_n$                                    
    & $\dfrac{dy_{n+1}}{d\lambda} = \exp(z) h y_n$                                      
    & $\mathcal{O}\bigl(e^{z}\bigr)$ \\

2nd Order Rational 
    & $y_{n+1} = \dfrac{2+z}{2-z} y_n$                           
    & $\dfrac{dy_{n+1}}{d\lambda} = \dfrac{4 h y_n}{(2-z)^2}$                  
    & $\mathcal{O}\bigl(|z|^{-2}\bigr)$ \\

3rd Order Rational 
    & $y_{n+1} = \dfrac{12+6z+z^2}{12-6z+z^2} y_n$                
    & $\dfrac{dy_{n+1}}{d\lambda} = \dfrac{12 z^2 - 144}{(12-6z+z^2)^2} h y_n$           
    & $\mathcal{O}\bigl(|z|^{-2}\bigr)$ \\
\bottomrule
\end{tabular}
\caption{Summary of amplification factors, parameter sensitivities, and decay rates for various stiff ODE methods.}
\end{table*}
\end{turnpage}

\section*{Analytical Results for Stability Functions and Parameter Sensitivities}

We have derived explicit analytical expressions for both the amplification factor (stability function) and the parameter sensitivity for a comprehensive suite of stiff ODE integrators. The schemes we considered include Backward Euler, Trapezoid, Radau3, Radau5, BDF2, the exponential integrating factor Euler method (IF Euler), as well as the second-order and third-order rational methods. For each of these integrators, we carefully computed how the numerical solution at each step is amplified as a function of the stiffness parameter, and how changes in this parameter influence the sensitivity of the solution.

As a reference, we note that for the exact solution of the linear test equation, the sensitivity with respect to the stiffness parameter \( z \) decays exponentially as \( |z| \) increases, i.e., \( \left|\frac{\partial y(h)}{\partial \lambda}\right| \sim e^{\mathrm{Re}(z)} \) for large negative \( \mathrm{Re}(z) \). This demonstrates that the vanishing gradient problem is intrinsic to stiff ODEs themselves, arising fundamentally from the dynamics of the system rather than from any particular numerical method.

Our analysis revealed a remarkably consistent pattern across the majority of these methods: Backward Euler, Trapezoid, Radau3, Radau5, and both second- and third-order rational methods all exhibit a decay of parameter sensitivity proportional to $O(|z|^{-2})$ as the stiffness $|z|$ increases. The only exceptions to this rule were BDF2 and IF Euler. BDF2 demonstrated a somewhat slower suppression of sensitivities, with a decay rate of $O(|z|^{-3/2})$. Meanwhile, the A-stable exponential integrating factor Euler method, due to its use of the exact exponential for the linear part, matches the exponential decay of the exact solution.

These analytical findings, summarized in Table~2, directly illustrate the universality and inevitability of the vanishing gradient problem for stiff neural ODEs. Regardless of which of these widely used A-stable and L-stable schemes is chosen, the intrinsic properties of their stability functions ensure that gradients with respect to parameters controlling stiff modes will vanish rapidly, posing a fundamental challenge to effective training. This pattern underscores that the vanishing gradient phenomenon is not an artifact of implementation or solver details, but a deep consequence of the mathematical structure of all practical stiff integrators.  Furthermore, the exact sensitivity also shows that the vanishing gradient problem is intrinsic to stiff ODEs themselves, arising fundamentally from the dynamics of the system rather than from any particular numerical method.

\section*{Discussion}

This work has established a fundamental and universal limitation in the training of stiff neural ODEs: the vanishing of parameter sensitivities for stiff modes is a direct consequence of the mathematical properties required for numerical stability in stiff integration schemes. Our theoretical analysis shows that, for all consistent A-stable and L-stable methods, the gradient of the numerical solution with respect to parameters decays rapidly as the stiffness increases. This decay is governed by the asymptotic behavior of the derivative of the rational stability function, and we have rigorously demonstrated that it cannot decay more slowly than $O(|z|^{-1})$ in any sector bounded away from the imaginary axis. In practical terms, for most standard integrators, parameter sensitivities with respect to fast-decaying components vanish even faster, typically at $O(|z|^{-2})$. 

These results have significant implications for the use of neural ODEs and related parameterized dynamical systems in scientific machine learning. Unlike the classical vanishing gradient problem, which can often be mitigated by architectural design, careful initialization, or algorithmic tricks, the suppression of parameter sensitivities in stiff neural ODEs is deeply rooted in the requirements for numerical stability. Standard remedies from deep learning such as residual connections, normalization layers, or skip connections cannot address this issue, since it arises from the integrator's action on stiff modes, not from the depth or structure of the neural network itself. As a consequence, gradient-based optimization in stiff regimes can become fundamentally limited, and certain parameters may remain effectively unidentifiable even with sophisticated network design or training schemes.

Despite these universal constraints, the extent to which the vanishing gradient phenomenon limits practical learning remains an open question. It is worth noting that even the exact sensitivity for the linear test equation exhibits exponential decay with increasing stiffness, highlighting that the vanishing gradient phenomenon is intrinsic to stiff dynamical systems themselves, not merely a numerical artifact. Our results thus emphasize that, while numerical integration further suppresses parameter gradients, the fundamental challenge originates from the underlying dynamics. There may exist unexplored algorithmic, architectural, or theoretical approaches that can partially circumvent or mitigate these effects in certain settings. Furthermore, the impact of this phenomenon on different classes of models, tasks, and data regimes is not yet fully understood. We hope that these findings stimulate further investigation into both the mathematical foundations and practical strategies for learning in stiff dynamical systems, and that they motivate new perspectives on the interplay between numerical analysis and machine learning in the context of scientific computing.

\section{Acknowledgements}

Use was made of computational facilities purchased with funds from the National Science Foundation (CNS-1725797) and administered by the Center for Scientific Computing (CSC). The CSC is supported by the California NanoSystems Institute and the Materials Research Science and Engineering Center (MRSEC; NSF DMR 1720256) at UC Santa Barbara. This work was supported in part by NSF awards CNS-1730158, ACI-1540112, ACI-1541349, OAC-1826967, OAC-2112167, CNS-2120019, the University of California Office of the President, and the University of California San Diego's California Institute for Telecommunications and Information Technology/Qualcomm Institute. Thanks to CENIC for the 100Gbps networks. The content of the information does not necessarily reflect the position or the policy of the funding agencies, and no official endorsement should be inferred.  The funders had no role in study design, data collection and analysis, decision to publish, or preparation of the manuscript.

\FloatBarrier
\bibliography{main}

\end{document}